\documentclass[10pt, fleqn, twocolumn]{article}

\usepackage[hang,flushmargin]{footmisc} 
\usepackage[utf8]{inputenc}
\DeclareUnicodeCharacter{2212}{$-$}
\usepackage{graphicx}
\usepackage[fleqn]{amsmath}
\usepackage[version=4]{mhchem}
\usepackage{siunitx}
\usepackage{longtable,tabularx}
\usepackage{nccmath}
\usepackage{amsmath}
\usepackage{array}
\usepackage{booktabs}
\usepackage[superscript]{cite}
\usepackage[margin=1in]{geometry}
\usepackage[toc,page]{appendix}
\usepackage{setspace}
\usepackage{titlesec} 
\usepackage{amssymb}
\usepackage{tabu}
\usepackage{listings}
\usepackage{afterpage}
\usepackage{titlesec}
\usepackage[labelfont=bf]{caption}
\usepackage{fancyhdr}
\usepackage{pgf}
\usepackage{float}
\usepackage{subcaption}
\usepackage{cleveref}
\usepackage{textcomp}
\usepackage{pict2e}
\usepackage{wasysym}
\usepackage{multicol}
\usepackage{lipsum}
\usepackage{indentfirst}
\usepackage{makecell} 

\usepackage[justification=centering]{caption}
\titleformat{\section}
{\normalfont\normalsize\bfseries}{\thesection}{0.5em}{}
\titleformat{\subsection}
{\normalfont\normalsize\bfseries\em}{\thesubsection}{0.5em}{}
\titleformat{\subsubsection}
{\normalfont\normalsize\bfseries\em}{\thesubsubsection}{0.5em}{}
\titlespacing*{\section}
{0pt}{3mm}{2mm}
\titlespacing*{\subsection}
{0pt}{2mm}{2mm}

\begin{document}

\twocolumn[
  \begin{@twocolumnfalse}

\begin{flushright}
\Large 

\end{flushright}
\begin{centering}      
\large 

\vspace{8mm}

\textbf{Optimizing Data Processing in Space for Object Detection in Satellite Imagery}\\
\vspace{0.5cm}
\normalsize 

Martina Lofqvist \& Jos\'e Cano\\
School of Computing Science, University of Glasgow\\
18 Lilybank Gardens, Glasgow, United Kingdom; +44 (0)1413301640\\
martinalofqvist@outlook.com\\


\vspace{0.5cm}
\centerline{\textbf{ABSTRACT}}
\vspace{0.3cm}
\end{centering}

With the cost of launch plummeting, it is now easier than ever to get a satellite to orbit. This has led to a proliferation of the number of satellites launched each year, resulting in downlinking of terabytes of data each day. The data received by ground stations is often unprocessed, making this an expensive process considering the large data sizes and that not all of the data is useful. This, coupled with the increasing demand for real-time data processing, has led to a growing need for on-orbit processing solutions. In this work, we investigate the performance of CNN-based object detectors on constrained devices by applying different image compression techniques to satellite data. We examine the capabilities of the NVIDIA Jetson Nano and NVIDIA Jetson AGX Xavier; low-power, high-performance computers, with integrated GPUs, small enough to fit on-board a nanosatellite. We take a closer look at object detection networks, including the Single Shot MultiBox Detector (SSD) and Region-based Fully Convolutional Network (R-FCN) models that are pre-trained on DOTA – a Large Scale Dataset for Object Detection in Aerial Images. The performance is measured in terms of execution time, memory consumption, and accuracy, and are compared against a baseline containing a server with two powerful GPUs. The results show that by applying image compression techniques, we are able to improve the execution time and memory consumption, achieving a fully runnable dataset. A lossless compression technique achieves roughly a 10\% reduction in execution time and about a 3\% reduction in memory consumption, with no impact on the accuracy. While a lossy compression technique improves the execution time by up to 144\% and the memory consumption is reduced by as much as 97\%. However, it has a significant impact on accuracy, varying depending on the compression ratio. Thus the application and ratio of these compression techniques may differ depending on the required level of accuracy for a particular task.\\

\\
  \end{@twocolumnfalse}
]


\section*{INTRODUCTION}

Over the past decade, we have seen a dramatic decrease in the cost of accessing space and a proliferation in the number of satellites in orbit. The introduction of nanosatellites, a small satellite with a wet mass between 1 to 10 kg, has increased the research and development in this field (Buchen and DePasquale, 2014). The use of nanosatellite missions for Earth Observation (EO) applications is expanding. EO applications include the monitoring and management of agriculture, water, forests, fisheries, and the climate. While the availability of data is increasing, there are significant challenges to the storage and transmission of the data to Earth. Terabytes of data are being transmitted to ground stations every day; an expensive process considering the large data sizes and that not all of the data is useful. Furthermore, the downlinking of data is restricted by the data rate and geographically limited to the locations of ground station networks. By developing space applications for on-orbit processing of raw data, we can minimize this bottleneck to save both time and money. Autonomous space systems offer many opportunities to continue exploring our solar system and beyond. 

The advancements in autonomous technologies and the resurgence of computer vision have led to a rise in demand for fast and reliable deep learning applications. However, these applications are mainly being developed for use on the ground, after the satellite data has been transmitted. Running Convolutional Neural Networks (CNNs) is a very computationally expensive process. It can be accelerated by using a Graphical Processing Unit (GPU) which exploits the potential for parallelism with many low power cores. In recent years, the industry has introduced small devices with impressive processing power to perform various autonomous processing tasks, such as increasingly powerful embedded GPUs including ARM Mali GPUs and NVIDIA Jetson boards. However, running CNNs on GPU devices still has limitations in terms of memory and compute capacity. In the space environment, resources are particularly limited. Therefore our objective is to investigate ways to improve the performance in terms of execution time and memory consumption. One approach can be to reduce the costs of the model, using a compression or optimization technique at some layer of the so-called Deep Learning Inference Stack (Turner, 2018). Another approach is to modify the input data.

This paper presents a new methodology for quantitatively assessing the performance of running object detection networks on satellite imagery aboard constrained devices. We evaluate different image compression techniques applied to satellite data to find a trade-off between execution time, memory consumption, and accuracy. This work builds on previous research in this area (Lofqvist et al., 2020), adding more realistic cases to the analysis, making a detailed comparison between lossy and lossless compression, and running the images on an additional device. The research aims to explore the advantages of computing at the edge in order to expedite the development of data processing in space for object detection in satellite imagery.

In the following sections, a brief background and related research are first discussed. Next, an overview of image compression techniques and its limitations are provided. Then, the approach is presented and a baseline for running the satellite data through object detection models is introduced. After this, the two compression techniques are evaluated against said baseline. Finally, we conclude and propose future work.


\section*{BACKGROUND \& PREVIOUS WORK}

\subsection*{Data Processing in Space}
Overall, deep learning applications performed on edge computing devices in space remain largely unexplored and only a few attempts to unveil its opportunities have been tested in orbit. Specifically, there has been little research conducted on running satellite data through neural networks on low-powered GPUs. Most of the research on deep learning applications and low power GPUs have been conducted for ground purposes and mainly investigate different hardware and software choices for improving performance on these constrained devices (Loukadakis et al., 2018; Rovder et al., 2019; Gibson, 2019; Radu et al., 2019). 

The processing environment for computers in space is radically different than on Earth (Tomayko, 1988). Denby and Lucia (2019) proposed an Orbital Edge Computer (OEC) system for implementing on orbit processing using a NVIDIA Jetson TX2 platform. Their work focused on nanosatellites with a solution for processing data locally when downlinking is not possible. They suggested that future work could investigate storage and machine-learning accelerators, providing context for our work. 

\subsection*{Object Detection on Satellite Imagery}
\begin{figure}[t]
    \centering
    \includegraphics[width=0.95\columnwidth]{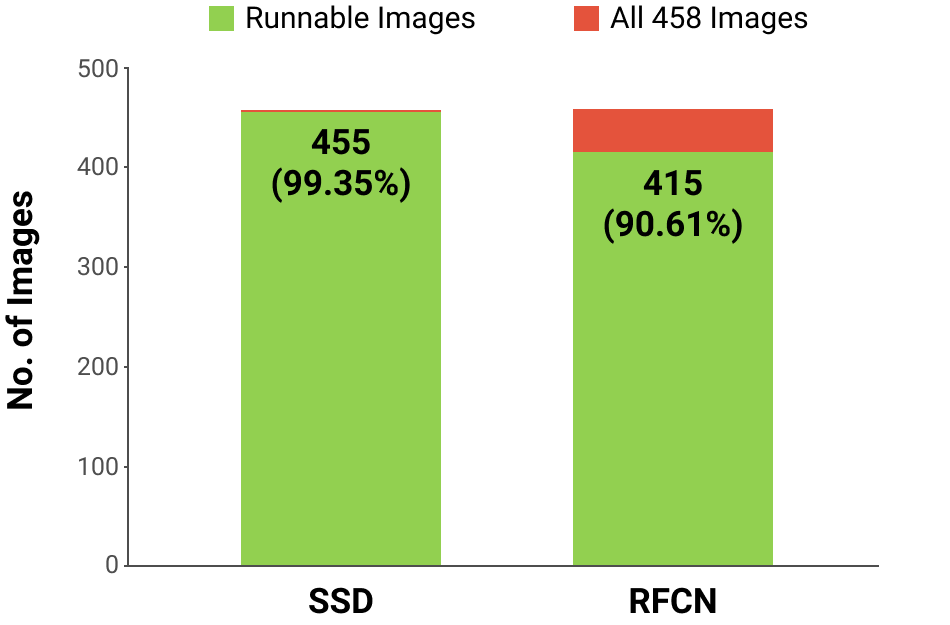}
    \caption{Runnable images of the DOTA dataset on the pre-trained SSD and R-FCN models on the NVIDIA Jetson Nano device. The figure shows that 99.35\% of the images can run on the SSD model and 90.61\% of the images can run on the R-FCN model}
    \label{fig:runnable}
\end{figure}

The quality and quantity of satellite images is improving the image processing task to characterize common objects on Earth's surface. The increasing importance of this task has led to the rapid development of new methods and applications for object detection in satellite images (Li et al., 2019). In previous work (Lofqvist et al., 2020), we investigated two models that can perform object detection on satellite images, pre-trained on the Dataset for Object Detection in Areal images (DOTA); Single Shot multibox Detector (SSD) and Region-based Fully Convolutional Network (R-FCN). This work showed that it was not possible to run the full dataset with the two models on a NVIDIA Jetson Nano device. The large satellite images in the dataset took a long time to process and sometimes consumed all of the available Random Access Memory (RAM) and SWAP memory that led the device to shut down. About 10\% of the images in the dataset on the R-FCN network were unable to run through the model (see Figure \ref{fig:runnable}). Applying compression techniques reduced the memory consumption in order to successfully run the full dataset on the Jetson Nano device. Additionally, three images were compared in terms of their performance after applying different compression ratios; the largest image, the smallest image, and the largest runnable image in the dataset. These images were of different sizes and aspect ratios, meaning that there was not a strong correlation amongst them to make a detailed comparison. Furthermore, it was found that the object detection models produced a higher accuracy for squared images. 

In this paper, the aim is to take a more realistic approach by investigating equally sized images in the dataset and making a detailed comparison between different compression techniques and constrained devices.

\section*{CHALLENGES OF REAL-TIME DATA PROCESSING IN SPACE}
Real-time data processing for object detection of satellite imagery faces three key challenges. First, physical constraints of the processors for space-based applications. Second, satellite data characteristics makes object detection a challenging task due to the lack of adequate datasets for training networks. Third, processing large satellite images on constrained devices requires resources that are not always available in the space environment. 

\subsection*{Resources in Space are Limited}
In the space environment, everything becomes more complex. Computers can show unexpected behaviors if the internal structures are changed due to the impacts of radiation (Wilson, George, and Klamm, 2016). Furthermore, deep learning applications are typically run on the cloud with powerful GPUs that speed up the processing. Since this is not accessible in space, memory- and power consumption and compute capacity of the device need to be accounted for. Additionally, the physical size of the processor matters and for the purpose of this work, it should be small enough to fit on board a nanosatellite. Smaller devices typically imply that there are fewer resources available, such as the number of cores and available memory. This has an impact on the number of operations the device is able to perform during a mission and the execution time for processing the images.

\subsection*{Lack of Adequate Training Data}
Object detection in satellite images is a complex task due to the inconsistencies of the backgrounds in these images, the low object-to-pixels ratios, the orientations of the objects to be detected, the variety of scales of the objects, and the low image resolutions. Therefore, classical object recognition algorithms often fail when applied to satellite imagery (Xia et al., 2018). There is a wealth of open source satellite imagery, however, there is a lack of robust datasets that contain annotated images and have an adequate amount of object categories, instances per image, and that come from a variety of optical sensors. This makes training of object detection networks a challenging task. 

\subsection*{Processing Large Satellite Data Drains Available Resources}
Satellite data is not only increasing in quantity, it is also increasing in size, with higher pixel dimensions and spatial resolutions. The level of detail of the images depends on the spatial resolution and is measured by Ground Sample Distance (GSD), which is influenced by the camera focal length, pixel sensor size, and orbital altitude. The quality of satellite images is continuously improving to extract more useful data. Today, companies are capturing satellite images with a spatial resolution of less than one meter\noindent{\footnote{https://www.planet.com/products/planet-imagery/}}.

Modern sensors on spacecraft are able to acquire up to a few thousand spectral bands. This requires large-scale on board storage systems that can store the data until the satellite passes a ground station for transmission. These high data volumes generated in space creates a bottleneck of transmitting the data to the ground. Furthermore, pre-processing these images to filter for useful data may consume additional memory and processing time. Therefore, there is an increasing need to compress the space data in order to effectively transmit it.


\section*{APPROACH}
In this section we discuss the design choices and implementations made to tackle the problem of running large images on constrained devices and we describe the compression techniques applied to the images to reduce the execution time and memory consumption. We choose to work with the DOTA dataset and use the pre-trained object detectors SSD and R-FCN. The edge devices used are the NVIDIA Jetson Nano and NVIDIA Jetson AGX Xavier with the Deep Learning framework TensorFlow\noindent{\footnote{https://www.tensorflow.org/}}. The results will help establish a trade-off between the time it takes to execute the task, the memory consumption, and the resulting accuracy.

\subsection*{Devices: NVIDIA Jetson Nano and NVIDIA Jetson AGX Xavier}
The most important factors to consider when determining a viable device for object detection aboard a nanosatellite are the volume, weight, and computing power. The device should be able to fit within a nanosatellite and run the images through a network without consuming all of the resources on board the satellite. Based on these factors, the NVIDIA Jetson Nano and the NVIDIA Jetson AGX Xavier devices are selected. Both devices include low-power, high capability, efficient GPUs, small enough to fit on-board a nanosatellite (Venturini, 2017). The NVIDIA Jetson platform powers a range of AI applications for computing at the edge and is compatible with many of the major deep learning frameworks.

\begin{figure}[h]
\centering
\begin{minipage}[b]{0.18\textwidth}
    \includegraphics[width=\textwidth]{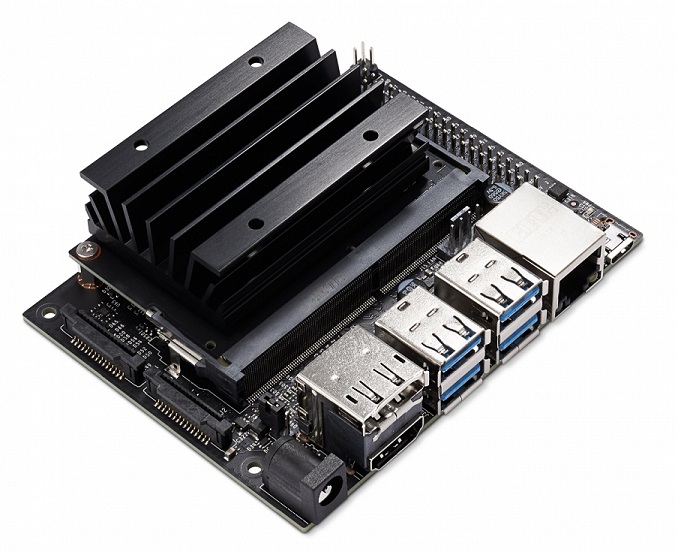}
    \label{fig:jetsonnano}
\end{minipage}
\begin{minipage}[b]{0.28\textwidth}
    \includegraphics[width=\textwidth]{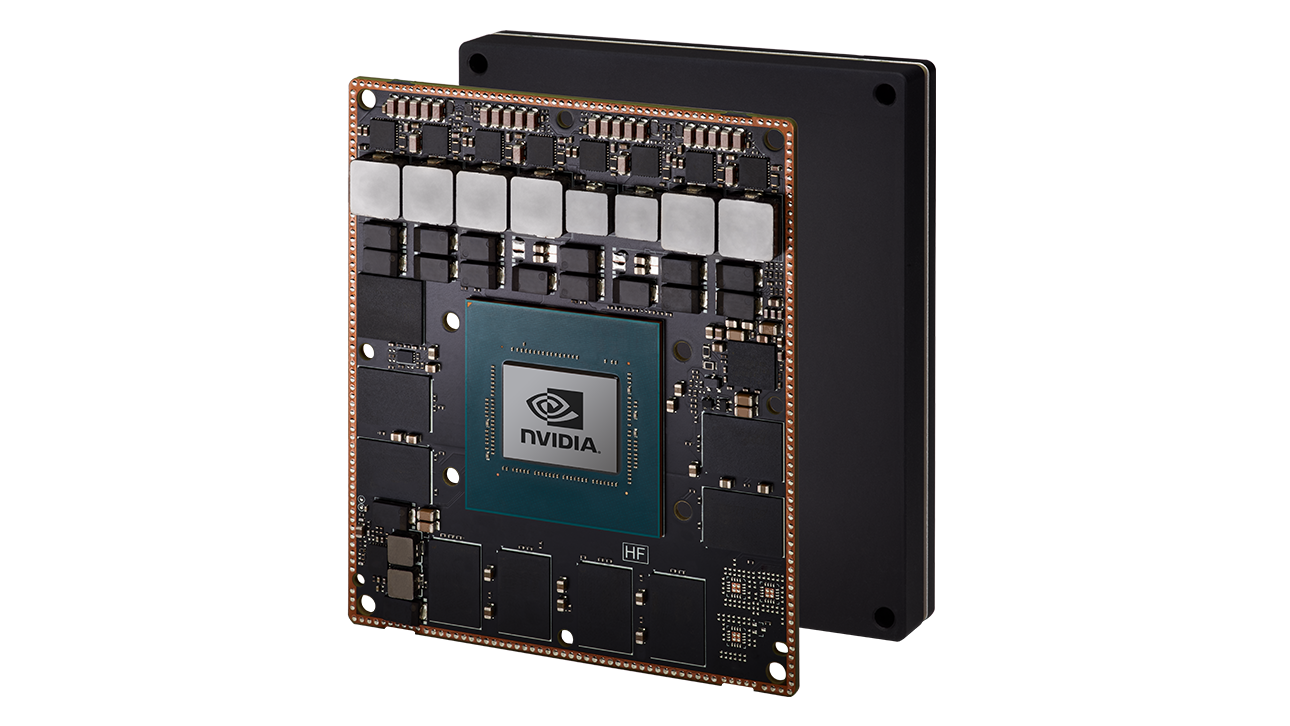}
    \label{fig:jetsonxavier}
\end{minipage}
\caption{NVIDIA Jetson Nano and AGX Xavier}
\label{fig:jetsons}
\end{figure}

The NVIDIA Jetson Nano\noindent{\footnote{https://www.nvidia.com/en-us/autonomous-machines/embedded-systems/jetson-nano/}} is 69.6 mm x 45 mm in size and comes with a 128-core integrated NVIDIA Maxwell GPU and a quad-core ARM CPU (Figure \ref{fig:jetsons}). It has 4GB of LPDDR4 memory at 25.6GB/s and 5W/10W power modes. 

Similarly, the NVIDIA Jetson AGX Xavier\noindent{\footnote{https://www.nvidia.com/en-us/autonomous-machines/embedded-systems/jetson-agx-xavier/}} is also a small device, 87 mm x 100 mm in size and comes with a 512-core NVIDIA Volta™ GPU with 64 Tensor Cores and an 8-core NVIDIA Carmel ARM CPU (Figure \ref{fig:jetsons}). It has 32GB of 256-bit LPDDR4x memory at 136.5GB/s and 10W/15W/30W power modes. 

\subsection*{Dataset: DOTA}
\begin{figure}[t]
    \centering
    \includegraphics[width = \columnwidth]{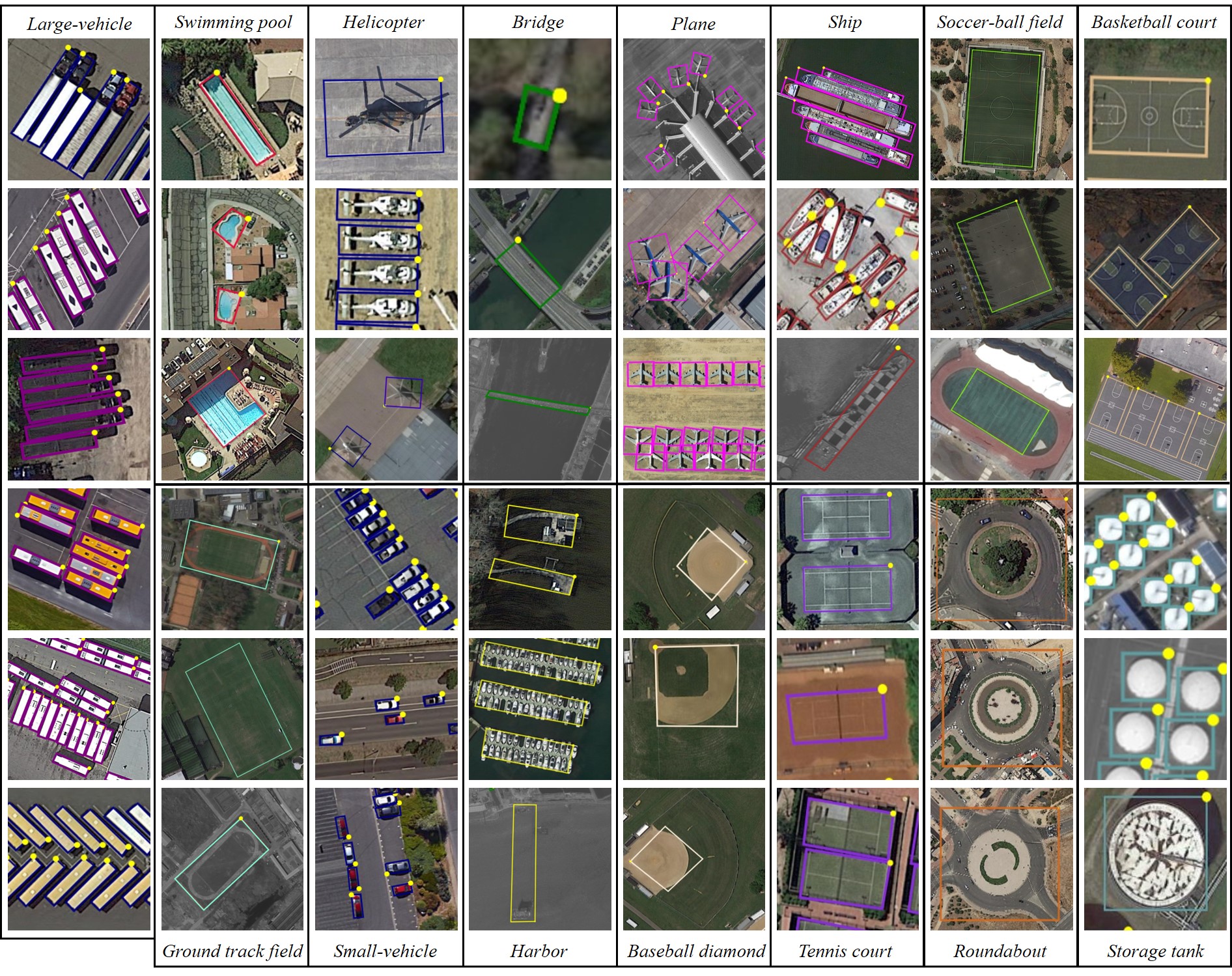}
    \caption{Samples of annotated instances and object categories in DOTA}
    \label{fig:dotainstances}
\end{figure}

\begin{figure}[h]
\centering
\begin{minipage}[b]{0.23\textwidth}
    \includegraphics[width=\textwidth]{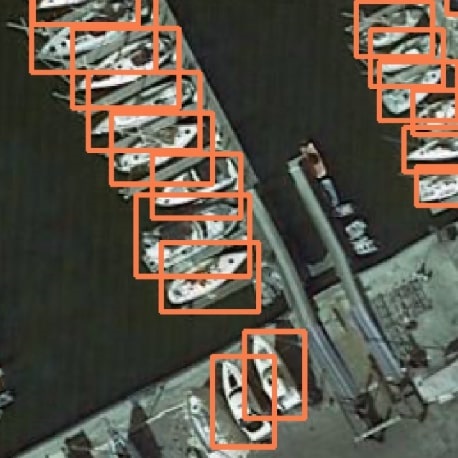}
    \label{fig:hbbships}
\end{minipage}
\begin{minipage}[b]{0.23\textwidth}
    \includegraphics[width=\textwidth]{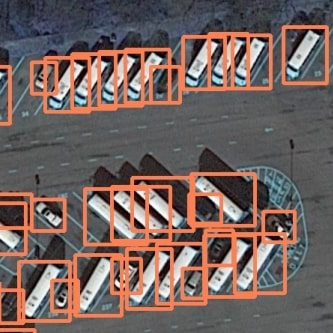}
    \label{fig:hbbvehicles}
\end{minipage}
\caption{Samples of crowded instances in annotated images in DOTA using horizontal bounding boxes (HBB)}
\label{fig:crowded-instances}
\end{figure}

The DOTA dataset contains satellite images from different sensors, with multiple annotated object instances, and realistic settings (Xia et al. 2018). A variety open-source models have been trained on this dataset with state-of-the-art object detection algorithms. The DOTA-v1.0 dataset consists of 2,806 images with a total of 188,282 annotated instances from 15 class categories, as seen in Figure \ref{fig:dotainstances}. The frequency of instances in each image varies with an average of about 67 object instances per image. Certain objects, such as cars and ships, are particularly difficult to detect as they often appear in high density areas. This makes it challenging to draw bounding boxes around those objects (see Figure \ref{fig:crowded-instances}). In contrast, large objects that do not appear in crowded areas, such as swimming pools and tennis courts, are often easier to detect. 

The images are collected from Google Earth, GF-2, and JL-1 satellites with resolutions ranging from 0.2 to 1 GSD. DOTA contains images with dimensions ranging from about $350\times 350$ to $7,500\times 7,500$ pixels. The images are in PNG format and represented as 24-bit true color (RGB). The dataset is split into training (50\%), validation (33.3\%), and testing (16.7\%). Both the training set and validation set have txt files available for each image containing the ground truth labels and object locations, the testing set does not. The instances are labelled with Horizontal Bounding Boxes (HBB) and Oriented Bounding Boxes (OBB). We analyze the performance of images from the validation set and we use the HBB ground truth labels since the pre-trained models provided by DOTA use these. The validation is split into two parts and we work with Part 1 that contains 458 images. 


\subsection*{Models: SSD \& R-FCN}
We work with the following models that are pre-trained on the DOTA dataset\noindent{\footnote{https://github.com/ringringyi/DOTA\_models}} using TensorFlow\noindent{\footnote{https://www.tensorflow.org/}} for object detection\noindent{\footnote{https://github.com/tensorﬂow/models/tree/master/\\research/object\_detection}}: 
\begin{itemize}
    \item \textbf{Single Shot MultiBox Detector (SSD)} with InceptionV2 backbone architecture
    \item \textbf{Region-based Fully Convolutional Network (R-FCN)} with ResNet-101 backbone architecture
\end{itemize}

\begin{figure}[t]
    \centering
    \includegraphics[width=0.95\columnwidth]{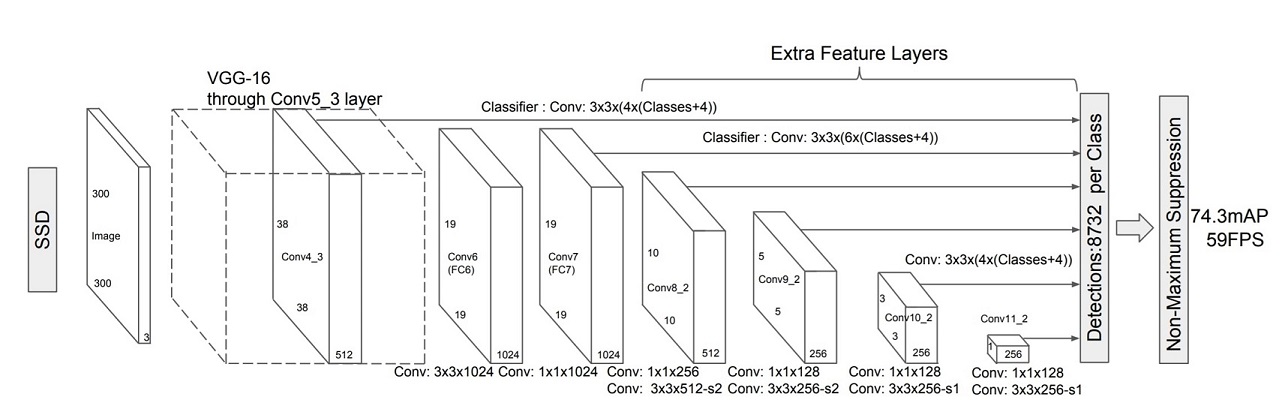}
    \caption{SSD Network Architecture}
    \label{fig:ssd}
\end{figure}

\textbf{SSD}, a fast single-shot object detector that predicts bounding boxes and confidence scores from a single pass (Liu, Anguelov, Erhan, Szegedy, Fu and Berg, 2016). It is a multibox detector with a single deep neural network, based on a feed-forward convolutional network (see Figure \ref{fig:ssd}). This means that the SSD is able to predict objects of various scales by combining different feature maps and default boundary boxes. Different layers handle the objects at a variety of scales, with the early network layers capturing more fine details of the input objects and are therefore used for high quality image classifications. However, compared to other detection methods, the SSD network generally performs worse on small objects (Zhao et al., 2019).

\begin{figure}[t]
    \centering
    \includegraphics[width=0.95\columnwidth]{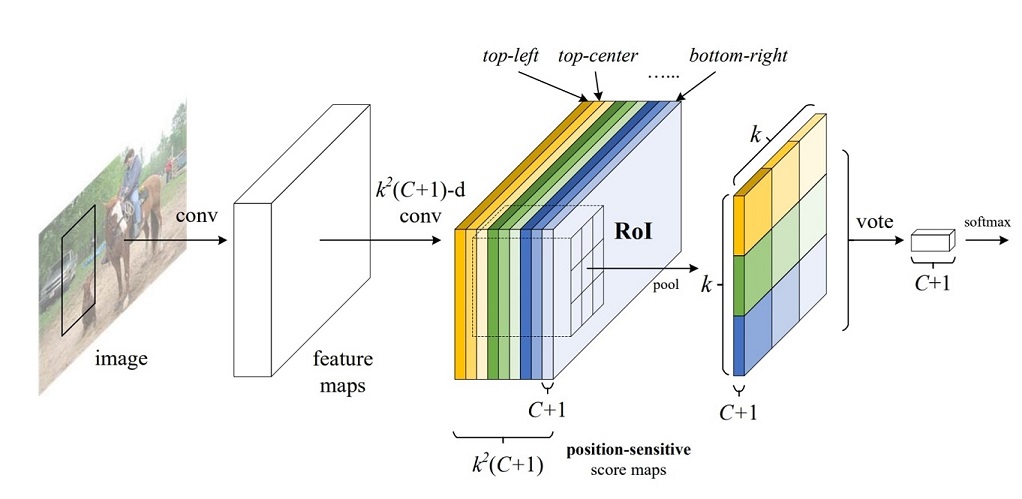}
    \caption{R-FCN Network Architecture}
    \label{fig:rfcn}
\end{figure}

\textbf{R-FCN} crops features from the last layer, prior to prediction, which means that per region computation is minimized which optimizes the speed. The cropping mechanism is location sensitive which improves the confidence score (Dai et al., 2016). Since the backbone architecture is the ResNet-101 model, R-FCN has 100 convolutional layers that are used to compute the feature maps, as seen in Figure \ref{fig:rfcn} (Dai et al., 2016).

\subsection*{Image Compression}
In order to run the full dataset on the Jetson devices, the images need to be reduced in size. We apply compression in order to achieve this. Compression techniques can be divided into two classes; lossless compression and lossy compression. The algorithms encode the information in fewer bits than the original image, which makes the file size smaller. All data can be recovered to its original format when applying the lossless compression method, while information is lost in the compression process when applying the lossy method.

\textbf{Lossless Compression}
Since there is no loss of information with this method, it is often used in applications that cannot tolerate any difference in information between the original and reconstructed data. This can be achieved using techniques such as Huffman coding and arithmetic coding (Bawa, 2010). Images with a larger redundancy can achieve a higher compression ratios. However, lossless compression techniques are generally unable to achieve high compression ratios. Optical satellite images usually have a lossless compression ratio of less than 3:1 (Qian, 2013). 


\textbf{Lossy Compression}
This compression technique is also known as image interpolation, which works by using existing data to predict values at unknown points. It tries to achieve the best approximation by using information from surrounding pixels. There are several interpolation algorithms that can be grouped into adaptive algorithms and non-adaptive algorithms (Kim et al., 2009). The adaptive algorithms change while the non-adaptive treat all pixels the same. Examples of adaptive algorithms include Bilinear and Nearest Neighbor (Kim et al., 2009). The Bilinear method provides better quality of the image, while the Nearest Neighbor is faster. The level of compression can be predetermined with a trade-off between file size and the speed of encoding/decoding. 


\subsection*{Our Implementation}
We apply the compression techniques to the satellite images prior to running them through the object detection networks to achieve the following goals:
\begin{itemize}
    \item To determine the necessary compression ratio in order to run the full dataset; and
    \item To optimize the performance in terms of execution time, memory consumption, and accuracy.
\end{itemize}

To achieve the first goal, we compress the largest image in the dataset by pixel dimension until it is able to run successfully on both models and Jetson devices. The SSD network is able to run larger images since it is less resource consuming than the R-FCN network with its additional number of convolutional layers. This analysis is performed on the Jetson Nano device since the Jetson AGX Xavier is capable of running all images. 

\begin{figure}[t]
    \centering
    \includegraphics[width = 6cm]{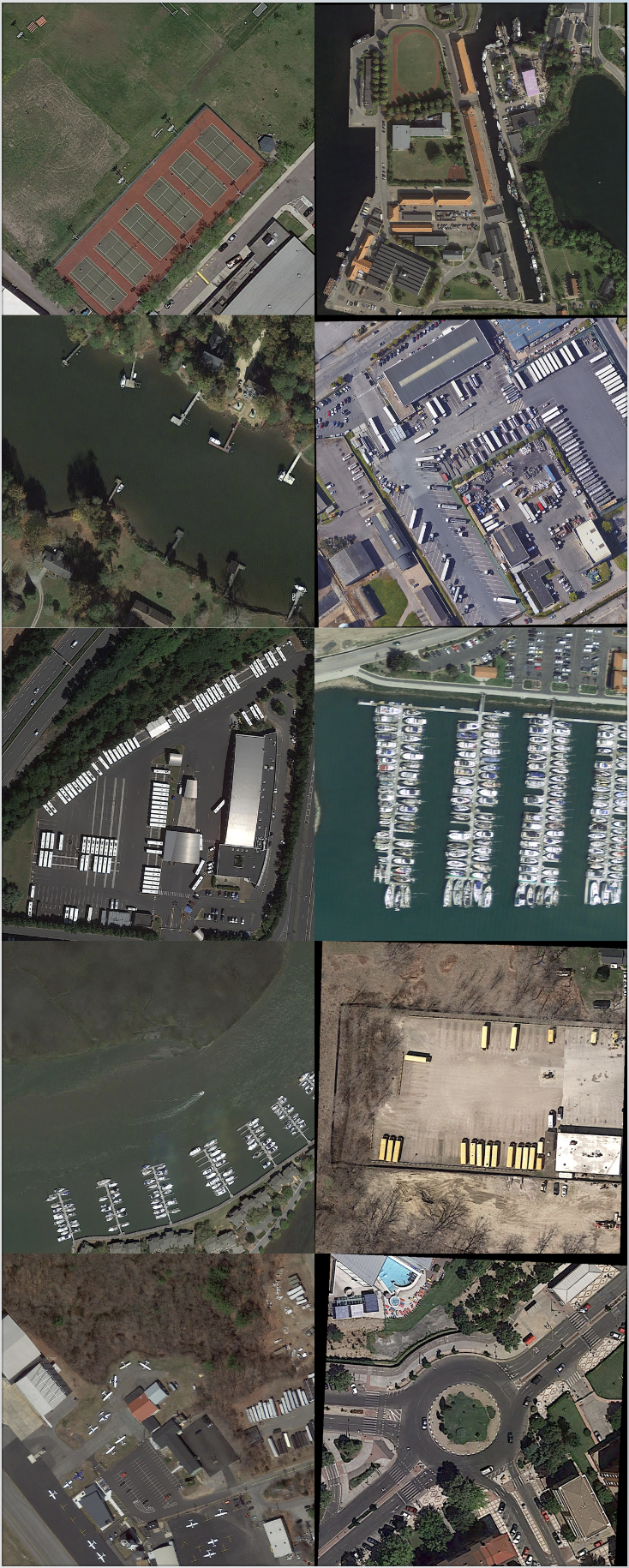}
    \caption{The set of 10 squared images from the DOTA validation set that will be evaluated}
    \label{fig:base_images}
\end{figure}

The second goal is achieved by first pre-processing the data to work with equally sized images. A random set of 10 images are selected from the validation set and cropped to squares with equal pixel dimensions of 843x843 pixels and sizes between 1MB and 1.5MB. The ground truth labels are edited accordingly with cropped object instances removed. The two compression techniques are evaluated using these 10 pre-processed images. The following images in Figure \ref{fig:base_images} will be assessed. 

These 10 images are then compressed using the lossless compression technique to as much as possible and to a 50\% and 10\% compression ratio with the lossy compression method. We then evaluate the performance and record the results. The execution time, memory consumption, and information about detected objects are given as output after running each image through the TensorFlow object detection code. The prediction labels are recorded with a confidence score, class category, and bounding box coordinates for each detected instance. The confidence score threshold is set to 0.01, meaning that we capture close to 100\% of all detected objects. Class categories are represented by an id from one of the 15 categories. The coordinates for each bounding box are returned in the format \texttt{(xmin, xmax, ymin, ymax)}. The methods used for evaluating the results are Average Precision (AP) and Intersection over Union (IoU). 



In terms of accuracy, the value is established by calculating the number of accurate objects detected with a confidence score of 10\% or more for the IoU score. Across all images, the R-FCN network showed higher accuracy than the SSD network. This is because SSD is a single stage detector, meaning that it only takes one pass on the image, while R-FCN is a multi-stage detector (Wu and Li, 2019). Therefore, higher accuracy of R-FCN and faster prediction of SSD are expected.


\section*{EVALUATION}

\subsection*{Baseline System}
The performance of the Jetson Nano and Jetson AGX Xavier are compared against a baseline. This baseline measures the inference time and memory consumption of the models by running them on a server with two GPUs; an NVIDIA Titan RTX and an NVIDIA Titan V. Table \ref{tab:hardware} shows some technical specifications of the baseline server and the two edge devices under study.

\begin{table}
\caption{Hardware Platforms}
\fontsize{7.5}{8.5}\selectfont
\begin{tabular}{|p{1.2cm}|p{1.6cm}|p{1.7cm}|p{1.6cm}|}
    \hline
     \textbf{} & \textbf{GPU \quad Memory} & \textbf{Cores} & \textbf{Memory Bandwidth}\\
    \hline
     \textbf{Titan RTX} & 24GB GDDR6 & \parbox[t]{1.7cm}{4608 CUDA,\\ 576 Tensor, \\72 RT} & 672 GB/s \\
    \hline 
      \textbf{Titan V} & 12GB HBM2 & \parbox[t]{1.7cm}{5120 CUDA,\\ 640 Tensor} & 652.8 GB/s \\
    \hline 
     \textbf{Jetson AGX Xavier} & 32GB LPDDR4x & \parbox[t]{1.7cm}{512 CUDA,\\ 64 Tensor} & 136.5 GB/s\\ 
    \hline
     \textbf{Jetson Nano} & 4GB LPDDR4 & 128 CUDA & 25.6 GB/s \\
    \hline
\end{tabular}
\label{tab:hardware}
\end{table}

\begin{figure}[t]
  \centering
  \includegraphics[width=\columnwidth]{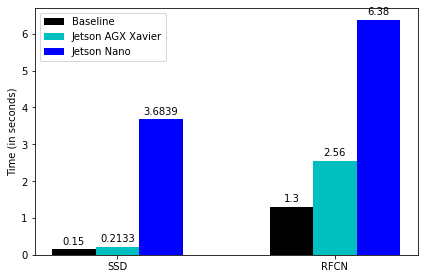}
\caption{Baseline Average Execution Time (in seconds) of the 10 squared images run through both models on the Jetson AGX Xavier and Jetson Nano devices, compared to a baseline.}
\label{fig:bas_inf}
\end{figure}

The set of 10 squared images run through the SSD and R-FCN models on the server and edge devices, and the inference time, memory consumption, and accuracy are recorded. The results show that it takes a much longer time to execute the process on the Jetson Nano and Jetson AGX Xavier devices compared to the server (see Figure \ref{fig:bas_inf}). On the Jetson AGX Xavier, it takes 1.5$\times$ longer for the SSD model and 2$\times$ for the R-FCN model to run compared to on the server. On the Jetson Nano, it takes 25$\times$ longer for the SSD model and 5$\times$ longer for the R-FCN model to run compared to on the server. This is a significant difference in execution time, so the aim is to reduce the time on the Jetson devices to get closer to the server results.  

We then apply compression techniques to these images to decrease the file size, thus consuming less memory and decreasing the inference time. We also assess the impact on accuracy. 

\subsection*{Running the Full Dataset}
The first task is to successfully run the full dataset. In order to run the full dataset, we resize the largest image, P1854, until it is capable of running on the slower R-FCN network. The execution time of the compression task itself is negligible. The results show that when the largest image is compressed to 59\% of its original size, with a pixel dimension of $7896 \times 2529$, the image is able to run successfully on both networks and on the Jetson Nano device. The average time for this resized image on the R-FCN network is 8.23 seconds, and 60 out of 221 objects (27\%) were detected accurately. The lossless compression technique was unable to compress this image to a small enough size for it to run successfully on both models. 

\subsection*{Results from Compressing the 10 Images}
The 10 images are compressed using the lossless method to the maximum amount possible and using the lossy method to 50\% and 10\% (see Table \ref{tab:specs}). 

\begin{table}
\caption{Original and Compressed sizes of the Images (in MBs)}
\fontsize{6.5}{5.5}\selectfont
\setlength{\aboverulesep}{2pt}
\setlength{\belowrulesep}{2pt}
\centering
\begin{tabular}{|c|c|c|c|c|}
    \toprule
     \textbf{} & \textbf{Original} & \textbf{Lossless} & \textbf{50\% Lossy} & \textbf{10\% Lossy}\\
     \midrule
     \textbf{P0104} & 1.34 & 1.32 & 0.4 & 0.02\\
     \midrule 
     \textbf{P0249} & 1.21 & 1.20 & 0.37 & 0.02\\
     \midrule
     \textbf{P0499} & 1.27 & 1.24 & 0.39 & 0.02\\ 
     \midrule
     \textbf{P0550} & 1.03 & 1.01 & 0.32 & 0.01\\
     \midrule
     \textbf{P0654} & 1.17 & 1.15 & 0.37 & 0.02\\
     \midrule
     \textbf{P0801} & 1.27 & 1.25 & 0.38 & 0.02\\
     \midrule
     \textbf{P1090} & 1.11 & 1.09 & 0.31 & 0.01\\
     \midrule
     \textbf{P1982} & 1.34 & 1.32 & 0.39 & 0.02\\
     \midrule
     \textbf{P2781} & 1.49 & 1.47 & 0.43 & 0.02\\
     \midrule
     \textbf{P2794} & 1.45 & 1.42 & 0.38 & 0.02\\
     \bottomrule
\end{tabular}
\label{tab:specs}
\end{table}

\textbf{Lossless Compression.}
The images compressed using the lossless compression had an improved execution time of about 10\% on both the Jetson AGX Xavier and the Jetson Nano for the SSD network and 8\% for the R-FCN network. The memory consumption was reduced to 3\% on the Jetson AGX Xavier and 4\% on the Jetson Nano for the SSD network and 2\% respectively 3\% for the R-FCN network. The accuracy remained the same since applying a lossless compression technique does not remove information from the image.

\begin{figure}[t]
\centering
\begin{subfigure}{.92\columnwidth}
  \centering
  \includegraphics[width=\columnwidth]{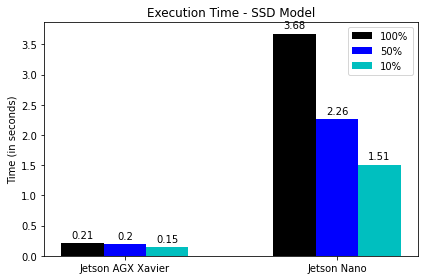}
  \subcaption{SSD}
  \label{fig:comp_speed_ssd}
\end{subfigure}%
\newline
\begin{subfigure}{.92\columnwidth}
  \centering
  \includegraphics[width=\columnwidth]{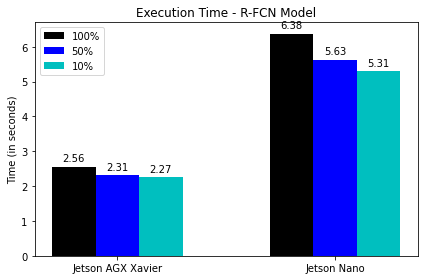}
  \subcaption{R-FCN}
  \label{fig:comp_speed_rfcn}
\end{subfigure}
\caption{Average Execution Time (in seconds) of running the images with the lossy compression technique with the two models, SSD and R-FCN, on the Jetson AGX Xavier and Jetson Nano devices. The execution time improves when applying the lossy compression technique}
\label{fig:comp_speed}
\end{figure}

\textbf{Lossy Compression.}
In the majority of cases, the speed was improved when the compression rate increased (see Figure \ref{fig:comp_speed}). On average, the time improved by 29\% on the Jetson AGX Xavier and 59\% on the Jetson Nano for the SSD model, and for the R-FCN model the time improved by 11\% on the Jetson AGX Xavier and 17\% on the Jetson Nano. These numbers are comparing the original image to the compressed image at a 10\% compression ratio. 

However, in most cases, the execution time for the compressed images did not match that of the baseline. The only case where the time did decrease sufficiently was for the 10\% compression ratio run on the SSD model on the Jetson AGX Xavier. Further compression would be necessary in order to reach the baseline results.

\begin{figure}[t]
\centering
\begin{subfigure}{.90\columnwidth}
  \centering
  \includegraphics[width=\columnwidth]{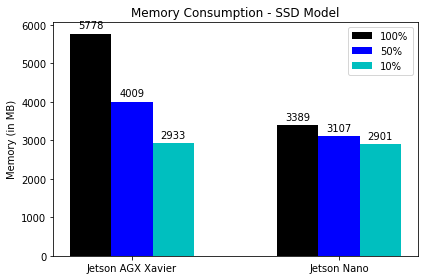}
  \subcaption{SSD}
   \label{fig:res_swap_ssd}
\end{subfigure}%
\newline
\begin{subfigure}{.90\columnwidth}
  \centering
  \includegraphics[width=\columnwidth]{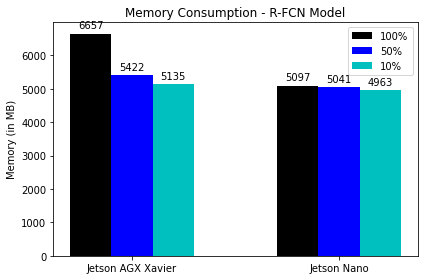}
  \subcaption{R-FCN}
  \label{fig:res_swap_rfcn}
\end{subfigure}
\caption{Average memory consumption (in MB) of running the images with the lossy compression technique with the two models, SSD and R-FCN, run on the Jetson Nano and Jetson AGX Xavier devices. The images improve in memory consumption when applying the lossy compression technique}
\label{fig:memory}
\end{figure}

The memory consumption also decreased for the compressed images (see Figure \ref{fig:memory}). The original image when compressed to 10\% and run through the SSD network on the Jetson AGX Xavier and Jetson Nano devices, saved 49\% and 14\% respectively. On the R-FCN network, the improvement was 23\% on the Jetson AGX Xavier and 3\% on the Jetson Nano. The correlation between inference time and memory consumption was expected as images with less pixels will be quicker to analyze and require less memory.

In terms of precision, objects that had a confidence score above the 10\% threshold were recorded. In the original image, the SSD network was able to detect 70 objects across the 10 images. This decreased to 61 objects (87\%) at a 50\% compression ratio and 8 objects (11\%) at a 10\% compression ratio. For the R-FCN network, 585 objects were detected across the 10 images. When this image was compressed to 50\%, 386 objects (66\%) were detected, and at 10\%, 86 objects (15\%) were detected. This was a decrease by almost 7$\times$ between the original image and the image with a 10\% compression ratio.

In terms of accuracy, the confidence scores between the original and the compressed images differed. In many cases, an object that had a confidence score of above 90\% in the original image would have a score of around 70\% in the images that were compressed to a 50\% compression ratio. 



\section*{CONCLUSION}

In this paper we build on previous research (Lofqvist, 2020) by investigating two pre-trained models, SSD and R-FCN, two devices, Jetson AGX Xavier and Jetson Nano, and two compression techniques, lossless and lossy image compression. These were evaluated using 10 images from the validation set in the DOTA dataset. The images were pre-processed in order for the set to contain equally sized and shaped images and a baseline was established by running these images on a powerful server. In comparison to the baseline, the Jetson AGX Xavier and Jetson Nano had a longer execution time and consumed more memory. Hence, lossless and lossy compression techniques were applied to improve the performance while recording the accuracy. We saw that the lossy compression technique was more effective in terms of improving the execution time and reducing the memory consumption. However, this technique compromised the precision and accuracy. Which compression technique to use depends on the application. If it is acceptable for the accuracy and precision to decrease, then the lossy compression technique may be considered. If not, then the lossless compression technique would be an alternative. 

Although the results on the edge devices were improved by compressing the images, they did not perform as well as the server. Therefore, some future work looking into ways to compress the models, SSD and R-FCN, could improve the performance further. Additionally, an alternative compression technique could be developed specifically to this cause.

\subsection*{References}
\begin{enumerate}

\itemsep0em 
    \item Bawa, S. (2010), ‘Compression Using Huffman Coding’, International Journal of Computer Science and Network Security, VOL.10 No.5. 
    \item Denby, B. and Lucia, B.(2019), ‘Orbital edge computing: Machine inference in space’, IEEE Computer Architecture Letters 18(1), 59–62.
    \item Buchen, E. and DePasquale, D. (2014), ‘Nano/ microsatellite market assessment’, SpaceWorks Enterprises. 
    \item Dai, J., Li, Y., He, K. and Jian, S. (2016), ‘R-FCN: object detection via region-based fully convolutional networks’, arXiv:1605.06409v2. 
    \item Gibson, P. (2019), ‘Deep learning on a low power gpu’, University of Edinburgh, Project Archive. 
    \item Kim, H., Park, S., Wang, J., Kim, Y. and Jeong, J.(2009), ‘Advanced bilinear image interpolation based on edge features’, IEEE.
    \item Li, K., Wan, G., Cheng, G., Meng, L., Han, J. (2019), Object Detection in Optical Remote Sensing Images: A Survey and A New Benchmark.
    \item Lofqvist, M., Cano, J. (2020), ‘Accelerating Deep Learning Applications in Space’, Workshop at 34th Annual Small Satellite Conference.
    \item Liu, W., Anguelov, D., Erhan, D., Szegedy, C., Fu, C. and Berg, A. C. (2016), ‘SSD: single shot multibox detector’, arXiv: 1512.02325v5. 
    \item Loukadakis, M., Cano, J. and O’Boyle, M. (2018), ‘Accelerating Deep Neural Networks on Low Power Heterogeneous Architectures’, 11th Int. Workshop on Programmability and Architectures for Heterogeneous Multicores. 
    \item Qian, S. (2013), Satellite Data Compression, Chapter 4.
    \item Radu V., Kaszyk K., Wen Y., Turner J., Cano J., Crowley E. J., Franke B., O'Boyle M., and Storkey A. (2019), Performance Aware Convolutional Neural Network Channel Pruning for Embedded GPUs, IEEE International Symposium on Workload Characterization (IISWC).
    \item Rovder, S., Cano, J. and O’Boyle, M. (2019), ‘Optimising convolutional neural networks inference on low-powered gpus’, 12th Int. Workshop on Programmability and Architectures for Heterogeneous Multicores. 
    \item Tomayko, J. E. (1988), Computers in Spaceﬂight: the NASA Experience, NASA Technical Reports Server. 
    \item Turner J., Cano J., Radu V., Crowley E. J., O'Boyle M., and Storkey A. (2018), Characterising Across-Stack Optimisations for Deep Convolutional Neural Networks, IEEE International Symposium on Workload Characterization (IISWC).
    \item Venturini, C. C. (2017), ‘Improving mission success of cubesats’, Aerospace Report, Space System Group. 
    \item Wilson C., George A., and Klamm B. (2016). 'A methodology for estimating reliability of SmallSat computers in radiation environments', 2016 IEEE Aerospace Conference, pp. 1-12.
    \item Wu, S. and Li, X. (2019), ‘Iou-balanced loss functions for single-stage object detection’, arXiv:1908.05641. 
    \item Xia G.-S., Bai X., Ding J., Zhu Z., Belongie S., Luo, J., Datcu M., Pelillo M. and Zhang L. (2018), ‘Dota: A large-scale dataset for object detection in aerial images’, IEEE Conference on Computer Vision and Pattern Recognition.
    \item Zhao, Z.-Q., Xu, S.-t. and Wu, X. (2019), ‘Object detection with deep learning: A review’, IEEE. arXiv:1807.05511v2. 
    \item Zhu H., Chen X., Dai W., Fu K., Ye Q., and Jiao J. (2015), ‘Orientation robust object detection in aerial images using deep convolutional neural network’, IEEE.
\end{enumerate}





\end{document}